\documentclass[conference]{IEEEtran}
\IEEEoverridecommandlockouts

\usepackage{cite}
\usepackage{color}
\usepackage{bm}

\usepackage{graphicx}
\makeatletter
\newcommand{\shorteq}{%
  \settowidth{\@tempdima}{-}
  \resizebox{\@tempdima}{\height}{=}%
}
\makeatother

\usepackage{amsmath}
\usepackage{amssymb}
\usepackage{subcaption}

\usepackage{hyperref} 

\usepackage{multirow}

\usepackage{tikz}
\usetikzlibrary{arrows} 
\usetikzlibrary{calc} 
\usetikzlibrary{arrows.meta}

\usepackage{pgfplots}

\usepackage{nicefrac} 

\usepackage[capitalize]{cleveref} 

\usepackage[normalem]{ulem}

\usepackage{comment}

\begin{document}

\title{Interaction and Decision Making-aware Motion Planning using Branch Model Predictive Control\thanks{This work was partially supported by the Wallenberg AI, Autonomous Systems and Software Program (WASP) funded by the Knut and Alice Wallenberg Foundation.}
}

\author{\IEEEauthorblockN{Rui Oliveira}
\IEEEauthorblockA{\textit{Division of Decision and Control Systems} \\
\textit{KTH Royal Institute of Technology}\\
Stockholm, Sweden \\
rfoli@kth.se}
\and
\IEEEauthorblockN{Siddharth H. Nair}
\IEEEauthorblockA{\textit{Model Predictive Control Laboratory} \\
\textit{UC Berkeley}\\
Berkeley, USA \\
siddharth\_nair@berkeley.edu}
\and
\IEEEauthorblockN{Bo Wahlberg}
\IEEEauthorblockA{\textit{Division of Decision and Control Systems} \\
\textit{KTH Royal Institute of Technology}\\
Stockholm, Sweden \\
bo@kth.se}
}

\maketitle

\begin{abstract}
Motion planning for autonomous vehicles sharing the road with human drivers remains challenging.
The difficulty arises from three challenging aspects: human drivers are 1) multi-modal, 2) interacting with the autonomous vehicle, and 3) actively making decisions based on the current state of the traffic scene.
We propose a motion planning framework based on Branch Model Predictive Control to deal with these challenges.
The multi-modality is addressed by considering multiple future outcomes associated with different decisions taken by the human driver.
The interactive nature of humans is considered by modeling them as reactive agents impacted by the actions of the autonomous vehicle.
Finally, we consider a model developed in human neuroscience studies as a possible way of encoding the decision making process of human drivers.
We present simulation results in various scenarios, showing the advantages of the proposed method and its ability to plan assertive maneuvers that convey intent to humans.
\end{abstract}

\def\RobotScript{A}
\def\HumanScript{H}

\def\sConflict{s_\text{conflict}}

\def\BranchingScript{\text{br}}
\def\sBranching{s_{\BranchingScript}}

\newcommand{\PolicySuperOrSubScript}{^}

\newcommand{\Policy}[1]{\pi \PolicySuperOrSubScript {#1}}
\newcommand{\ProbPolicy}[1]{P_{ \Policy{#1} }}

\def\inputAV{u^{\RobotScript}}
\def\inputHV{u^{\HumanScript}}

\def\policyBefore{\pi \PolicySuperOrSubScript \text{before}}
\def\policyAfter{\pi \PolicySuperOrSubScript \text{after}}

\def\pathLengthHV{s^{\HumanScript}}
\def\pathLengthAV{s^{\RobotScript}}
\def\velocityHV{v^{\HumanScript}}
\def\velocityAV{v^{\RobotScript}}

\def\velocityRefHV{v_\text{ref}}

\def\policyVelocityTracking{\pi \PolicySuperOrSubScript {\velocityRefHV,\text{track}}}
\def\policyVehicleAhead{\pi \PolicySuperOrSubScript {\text{va}}}

\def\dLeading{d}
\def\dLeadingRef{d_{\text{ref}}}

\def\velocityRefHVFast{v_\text{ref}^\text{fast}}
\def\velocityRefHVKeep{v_\text{ref}^\text{keep}}
\def\velocityRefHVSlow{v_\text{ref}^\text{slow}}

\def\policyVehicleAheadFast{\pi \PolicySuperOrSubScript {\text{va,fast}}}
\def\policyVehicleAheadKeep{\pi \PolicySuperOrSubScript {\text{va,keep}}}
\def\policyVehicleAheadSlow{\pi \PolicySuperOrSubScript {\text{va,slow}}}

\def\PolicyCross{\Policy{\text{cross}}}
\def\PolicyStop{\Policy{\text{stop}}}
\def\ProbPolicyCross{\ProbPolicy{\text{cross}}}
\def\ProbPolicyStop{\ProbPolicy{\text{stop}}}

\def\DeltaT{\Delta_{T}}

\def\GivenCurrentInformation{}

\def\SetOfNodes{\mathcal{J}}
\def\ActualNode{j'}
\def\SetOfNodesMinusRoot{\SetOfNodes \setminus \{0\}}
\newcommand{\SetOfNodesMinusArgument}[1]{\SetOfNodes \setminus \{{#1}\} }

\def\branchingTimeIndex{t_{\BranchingScript}}
\def\branchingState{x_{\branchingTimeIndex}^{\HumanScript,\pi \PolicySuperOrSubScript 0}}
\def\ObservationTime{\Delta t_{\text{obs}}}
\def\ObservationTimeInstant{\branchingTimeIndex + \ObservationTime}

\def\NodeHVStateVector{\bm{x}^{\HumanScript}_j}
\def\NodeAVStateVector{\bm{x}^A_j}

\newcommand{\NodeAVInputVector}[1][j]{\bm{u}^A_{#1}}
\newcommand{\NodeAVInputVectorOptimal}[1][j]{{\bm{u}^{A}_{#1}}^{\star}}

\def\InitialStateHV{x_{\text{init}}^{\HumanScript}}
\def\InitialStateAV{x_{\text{init}}^{\RobotScript}}

\def\PolicyRedLight{\Policy{\text{red}}}
\def\PolicyGreenLight{\Policy{\text{green}}}
\def\ProbPolicyRedLight{\ProbPolicy{\text{red}}}
\def\ProbPolicyGreenLight{\ProbPolicy{\text{green}}}

\def\ie{\textit{i.e.}}

\def\baselabel{}
\def\basepath{tex}
\def\basepathfigures{figures}
\def\basepathtikz {figures/tikz}


\section{Introduction}

Autonomous vehicles (AVs) must drive in the presence of other traffic participants, such as human-driven vehicles (HVs) and pedestrians.
To this date, sharing the road with human traffic participants is one of the biggest challenges hindering AVs from being deployed at a large scale.

Current state-of-the-art sensor and perception technology already provides an accurate understanding of the traffic scene's current state.
However, the irregularity of human behavior makes the prediction task, \textit{i.e.}, knowing how the traffic scene evolves, reliable only for a few seconds into the future.

Besides being hard to predict, the traffic scene evolution is also directly impacted by the decisions taken by the AV.
Human drivers will react differently depending on other surrounding vehicles' maneuvers.

Most planning approaches assume that other traffic participant predictions are fixed, resulting in the autonomous vehicle performing maneuvers to avoid said predictions.
The result is overly conservative driving from the autonomous vehicle.
To consider that the AV's decisions impact the traffic scene evolution and avoid conservativeness, one must solve the joint prediction and planning problem.

This work presents a novel motion planning approach to tackle the joint prediction and planning problem, making the following contributions:
\begin{itemize}
\item Proposal of a framework for handling interaction-heavy scenarios, considering the aspects of \emph{multi-modality}, \emph{interaction}, and \emph{decision making} of human drivers;
\item Approximating human drivers' decision making process through models developed in neuroscience studies, allowing the autonomous vehicle to take proactive and assertive maneuvers that convey intent to human drivers;
\item Performance evaluation and comparison against relevant works in the area, showing an increase in average performance without sacrificing safety.
\end{itemize}

\def\basepathtikzchallenges{figures/tikz-scenario}

\def\ScalingTikz{0.1}
\def\ScalingCar{0.09}
\def\DefaultVelocity{v_{r}}

\begin{figure*}
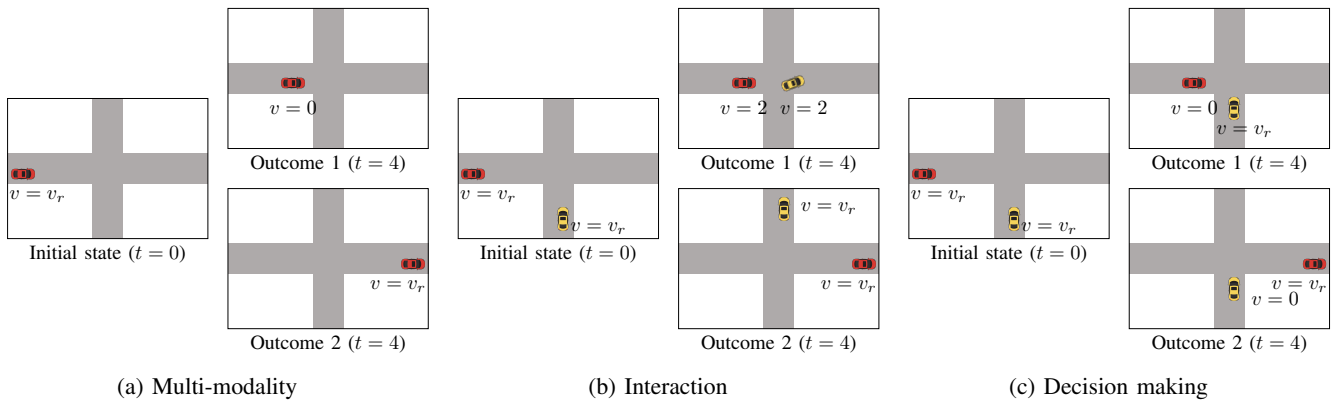

    \centering
    \begin{subfigure}[b]{0.31\textwidth}
    \centering
    \resizebox{1.05\textwidth}{!}{
    \begin{tikzpicture}[scale=1.75]
    
    \input{\basepathtikzchallenges/challenge-1}

    \end{tikzpicture}
    }
        \caption{Multi-modality}
        \label{\baselabel fig:multi-modal}
    \end{subfigure}    
    ~ 
    \begin{subfigure}[b]{0.31\textwidth}
    \centering
    \resizebox{1.05\textwidth}{!}{
    \begin{tikzpicture}[scale=1.75]
    
    \input{\basepathtikzchallenges/challenge-2}

    \end{tikzpicture}
    }
        \caption{Interaction}
        \label{\baselabel fig:interaction}
    \end{subfigure}
    ~ 
    \begin{subfigure}[b]{0.31\textwidth}
    \centering
    \resizebox{1.05\textwidth}{!}{
    \begin{tikzpicture}[scale=1.75]
    
    \input{\basepathtikzchallenges/challenge-3}

    \end{tikzpicture}
    }
        \caption{Decision making}
        \label{\baselabel fig:decision-making}
    \end{subfigure}
    \caption{Challenges associated with an autonomous vehicle (yellow car) driving in the presence of human drivers (red car).}\label{fig:challenges}
\end{figure*}

\subsection{Related work}
\label{sec:related_work}

We start by introducing three challenging aspects of human drivers and existing works relating to them.

\subsubsection{Multi-modality}

Consider a human driver arriving at an intersection (\emph{initial state} of~\cref{\baselabel fig:multi-modal}).
A defensive driver slows down and stops to check for oncoming vehicles safely (\emph{outcome 1}).
On the other hand, an aggressive driver speeds through the intersection to achieve a shorter traveling time (\emph{outcome 2}).
A planner must consider different outcomes and plan a motion that guarantees safety for all.

Model Predictive Control (MPC) approaches for tackling uncertainties stemming from the multi-modality are introduced in~\cite{batkovic2020robust,phiquepal2021control}.
However, they are limited to uncertainties stemming from the unknown existence of static obstacles or intents of pedestrians.

The works~\cite{schildbach2015scenario,cesari2017scenario} use Scenario MPC to consider the multi-modality arising from the uncertainty over different maneuvers types of other drivers (such as keep or change lanes).
\cite{brudigam2018scsmpc} combines Scenario MPC and Stochastic MPC to improve other vehicles' motion predictions.

Contingency MPC is an approach that tracks a desired nominal plan while maintaining a contingency plan to deal with possible emergencies~\cite{alsterda2019contingency,alsterda2021contingency}.
Multi-modality is tackled by considering both the nominal and contingency outcomes.

\subsubsection{Interaction}

Driving is a highly interactive task, where drivers adapt their actions in response to those of other drivers.
Consider the HV (red car) and AV (yellow car) approaching the intersection shown in \emph{initial state} of~\cref{\baselabel fig:interaction}.
If the AV turns right, the HV will slow down to avoid a collision (\emph{outcome 1}).
On the other hand, if the AV proceeds through the intersection, the HV keeps its speed (\emph{outcome 2}).
Interactions occur in most traffic situations, and considering them is crucial to reducing AVs' conservativeness.

The problem of driving an autonomous vehicle through an intersection is tackled with Stochastic MPC in~\cite{nair2022interaction}.
The approach considers the interaction aspects of driving by modeling human drivers as closed-loop predictions tracking a constant headway to the AV in front of them.

The work in~\cite{sadigh2018planning} introduces a formulation of interaction with HVs as an underactuated dynamical system.
This formulation allows the AV to perform complex interaction behaviors, such as accelerating or slowing down, to show intent to humans.

\subsubsection{Decision making}

Humans often make decisions while driving, as shown in the initial state of~\cref{\baselabel fig:decision-making}), where the HV (red car) has to decide if it will cross the intersection.
The decision is affected by the perceived intended behavior of the AV (yellow car).
If the AV approaches at high speed (\emph{outcome 1}), the human decides to stop at the intersection.
On the other hand, if the AV slows down (\emph{outcome 2}), the human drives through the intersection.
This example shows how the traffic scene, namely the AV's position and velocity, affects the HV's decision~\cite{marti2015drivers}.
This example significantly differs from the interaction aspects shown in~\cref{\baselabel fig:interaction}, as no imminent collision forces the HV to stop.
Instead, the AV shows an intent to drive through or stop at the intersection, leading the HV to react accordingly.

Branch MPC~\cite{chen2021interactive} is used to tackle the multi-modality arising from other human driver's decision making.
The human is modeled using a finite set of policies that build a scenario tree.
The planned solution is a feedback policy in the form of a trajectory tree accounting for all possible scenarios.
The HV policies are propagated independently of other agents, which can lead to the freezing robot problem~\cite{trautman2010unfreezing}.
Realizing this drawback, the authors of~\cite{wang2022interaction} consider closed-loop models to propagate the control policies of other vehicles.
However, both works lack a driver decision making model based on human behavior research.

\subsubsection{Summary}

\Cref{\baselabel table:approaches_comparison} outlines the previous works regarding their ability to tackle the mentioned challenges.
Only recently has~\cite{wang2022interaction} considered all challenges.
We build upon~\cite{chen2021interactive} by considering the reactive behavior of humans and addressing the \emph{interaction} challenge.
Moreover, we model human \emph{decision making} using neuroscience studies, adding a sound sociological model lacking in~\cite{wang2022interaction}.

\def\TableYes{\textcolor{green}{\checkmark}}
\def\TableNot{\textcolor{red}{$\times$}}

\newcommand{\TableCiteBatkovic}{\cite{batkovic2020robust}}
\newcommand{\TableCiteCesari}{\cite{cesari2017scenario}}
\newcommand{\TableCiteChen}{\cite{chen2021interactive}}
\newcommand{\TableCiteSadigh}{\cite{sadigh2018planning}}
\newcommand{\TableCiteAlsterda}{\cite{alsterda2021contingency}}
\newcommand{\TableCiteNairInteraction}{\cite{nair2022interaction}}
\newcommand{\TableCiteWang}{\cite{wang2022interaction}}

\begin{table}[ht]
\centering
\caption{Comparison of different planning approaches according to the challenges identified in~\cref{sec:related_work}. }
\begin{tabular}{l c c c}
Approach                     & Multi-modal & Interaction & Decision making \\
\hline
\TableCiteBatkovic ,\cite{phiquepal2021control},\TableCiteCesari ,\TableCiteAlsterda & \TableYes & \TableNot & \TableNot  \\
\TableCiteChen               & \TableYes & \TableNot & \TableYes  \\
\TableCiteSadigh             & \TableNot & \TableYes & \TableYes  \\
\TableCiteNairInteraction    & \TableYes & \TableYes & \TableNot  \\
\TableCiteWang , our approach                 & \TableYes & \TableYes & \TableYes  \\
\end{tabular}
\label{\baselabel table:approaches_comparison}
\end{table}

\section{Modeling}
\label{\baselabel sec:modeling}

\subsection{Scenarios}
\label{\baselabel subsec:modeling-scenarios}

\begin{figure}
    \centering
    \begin{subfigure}[b]{\columnwidth}
	    \centering
        \includegraphics[width=0.8\textwidth]{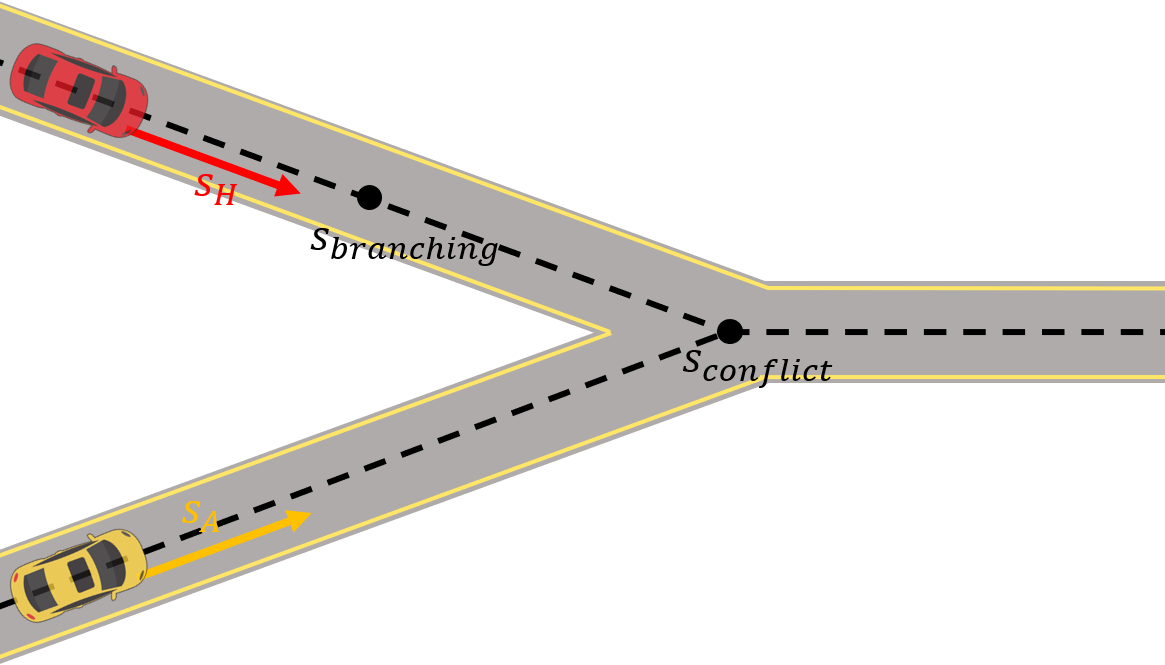}
        \caption{Merging scenario.
The AV (yellow) needs to merge into a lane where an HV (red) is also merging into.}
        \label{\baselabel fig:y_merging_scenario}
    \end{subfigure}
    \par\medskip
    
    \begin{subfigure}[b]{\columnwidth}
	    \centering
        \includegraphics[width=0.8\textwidth]{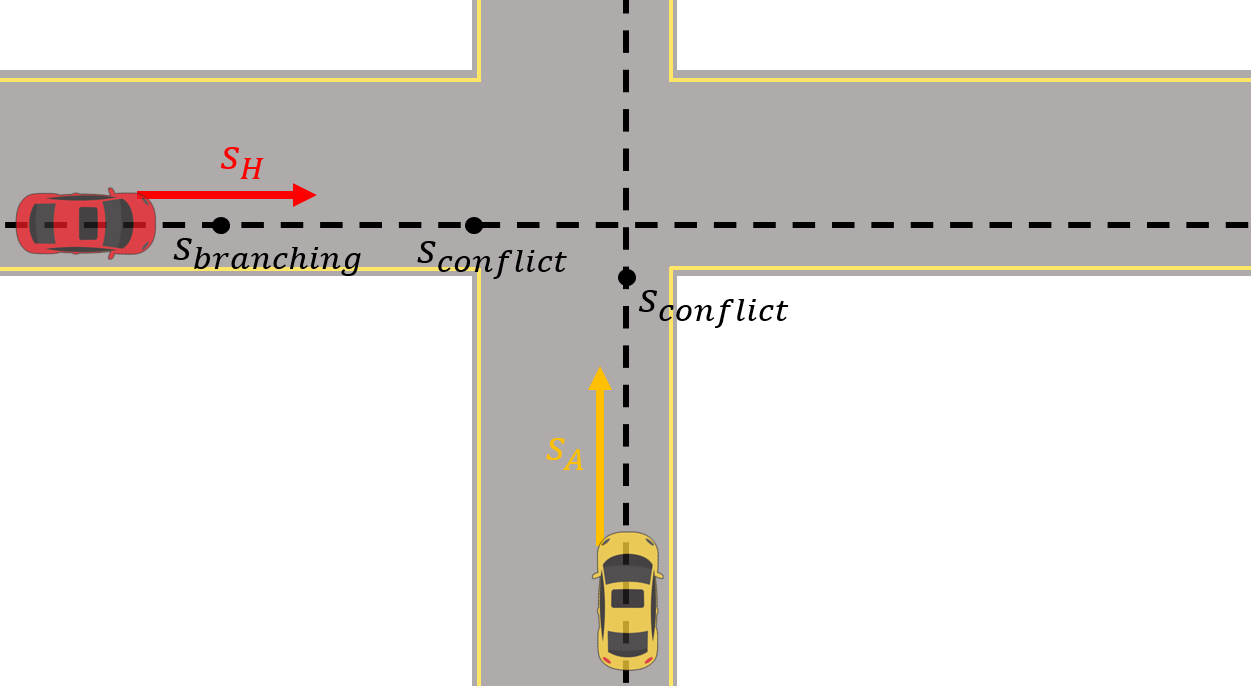}
        \caption{Intersection scenario.
The AV (yellow) and the HV (red) need to ensure that they do not cross simultaneously.}
        \label{\baselabel fig:intersection_scenario}
    \end{subfigure}
    \caption{The scenarios considered in this work.\label{fig:scenarios}}
\end{figure}

We consider two scenarios that force the AV to interact with a HV.
In the first scenario, the AV merges onto a road, as shown in~\cref{\baselabel fig:y_merging_scenario}.
The second scenario considers a non-signalized intersection, \ie, an intersection without traffic lights~\cref{\baselabel fig:intersection_scenario}.
Both scenarios lack priority rules, and thus no vehicle has the right of way over the other, requiring interaction and unspoken negotiation between them.

We assume that a vehicle $i$, $i \in \{\HumanScript, \RobotScript \}$, human-driven ($\HumanScript$) or autonomous ($\RobotScript$), moves along the road centerline along a path length $s^i$.
In both scenarios in~\cref{\baselabel fig:y_merging_scenario,\baselabel fig:intersection_scenario}, the vehicles are on separate lanes until the merging point $\sConflict$.
After $\sConflict$ the vehicles are on the same lane (merging), or share a conflict region (intersection), and must keep a safe distance to avoid a collision.
We assume that as the HV approaches the conflict point, it eventually makes a decision at $\sBranching$, possibly changing its behavior.
$\sBranching$ is a point located before $\sConflict$, and it is chosen so as to represent a point on the road where the HV becomes aware of the AV, and takes a decision according to its preferences and driving style,~\cite{schwarting2019social}, and even the intention shown by the AV~\cite{marti2015drivers}.

\subsection{Vehicle models}

We assume the AV to drive along the lane center and only plan for its longitudinal motion.
This assumption is justified as there is little to no advantage in considering the possibility of steering the vehicle laterally in these scenarios.
Both vehicles follow the model 
\begin{equation}
\label{\baselabel eq:vehicle_model}
x =
\begin{bmatrix}
s & v
\end{bmatrix}^\intercal,
\quad
\dot{x} =
\begin{bmatrix}
v & u
\end{bmatrix}^\intercal,
\end{equation}
where $s$ is the current position along the centerline path, $v$ is the vehicle longitudinal velocity, and $u$ is the acceleration, corresponding to the control input.

The AV input $\inputAV$ is determined by the planning framework presented in~\cref{\baselabel sec:mpc_formulation}.
The HV input $\inputHV$ follows a certain driving policy $\policyBefore$, up to $\sBranching$, and afterwards $\policyAfter$:
\begin{equation}
\label{\baselabel eq:policies_before_after_branching}
\inputHV = 
  \begin{cases} 
   \policyBefore(\pathLengthHV, \velocityHV) & \text{if } \pathLengthHV < \sBranching \\
   \policyAfter(\pathLengthHV, \velocityHV, \pathLengthAV, \velocityAV)  & \text{if } \pathLengthHV \geq \sBranching
  \end{cases}
\end{equation}
Note that $\policyAfter$ is a function of $\pathLengthAV$ and $\velocityAV$, due to the interaction behavior between the HV and AV.

\subsection{Human driving policies}
\label{\baselabel subsec:human_driving_policies}

We assume two types of policies for the HV, \emph{velocity tracking} when there is no vehicle ahead, and \emph{vehicle following} when there is a vehicle ahead.

\subsubsection{Velocity tracking}

When there is no leading vehicle ahead, the HV will track a desired reference speed $\velocityRefHV$.
We then have $\inputHV = \policyVelocityTracking$, where
\begin{equation}
\label{\baselabel eq:policy_velocity_tracking}
\policyVelocityTracking \left( \velocityHV, \velocityRefHV \right) = K_v \left(\velocityRefHV - \velocityHV \right),
\end{equation}
and $K_v > 0$ is a constant gain.
Policy $\policyVelocityTracking$ takes as inputs the current vehicle velocity $\velocityHV$ and the desired reference velocity $\velocityRefHV$, and outputs an acceleration command proportional to their difference.

\subsubsection{Vehicle following}
\label{\baselabel subsubsec:vehicle_following}

If there is another vehicle ahead, the HV adapts its speed to avoid a rear-end collision.
In this case, $\inputHV = \policyVehicleAhead$, where $\text{va}$ stands for \emph{vehicle ahead}, and:
\begin{equation}
\label{\baselabel eq:policy_velocity_following}
\policyVehicleAhead = 
\begin{cases} 
\policyVelocityTracking \left( \velocityHV, \velocityRefHV \right)   & \text{if } \velocityHV \geq \velocityRefHV \\
K_v (\velocityAV - \velocityHV) + K_d \left( \dLeading - \dLeadingRef \right) & \text{if } \velocityHV < \velocityRefHV
\end{cases}
\end{equation}
where $\dLeading = \pathLengthAV - \pathLengthHV$ corresponds to the distance from the human-driven vehicle to the vehicle ahead, and $K_d > 0$ is a constant gain.

When $\velocityHV \geq \velocityRefHV$, the vehicle tracks its desired velocity $\velocityRefHV$, resulting in braking.
When $\velocityHV < \velocityRefHV$, the vehicle speeds up while taking into account the vehicle ahead.
The term $K_v (\velocityAV - \velocityHV)$ performs velocity tracking, and $K_d \left( \dLeading - \dLeadingRef \right)$ keeps a safe distance $\dLeadingRef$ to the vehicle in front.
Policy $\policyVehicleAhead$ is inspired by the Intelligent Driver Model~\cite{treiber2000congested}.

\subsection{Human decision making}
\label{\baselabel subsec:human_decision_making}

In the considered scenarios we assume the human has two different behaviors, one before arriving at $\sBranching$, and another after it, as given by~\Cref{\baselabel eq:policies_before_after_branching}.
At $\sBranching$ the HV starts interacting with the AV and changes its behavior.
The new behavior depends on the type of driver profile and on the traffic scene, and it is not known in advance.
In the merge scenario we consider behaviors corresponding to three types of driver:
\begin{itemize}
\item $\policyVehicleAheadFast$ - egoistic driver who speeds up ($\velocityRefHV = \velocityRefHVFast$),
\item $\policyVehicleAheadKeep$ - neutral driver who keeps speed ($\velocityRefHV = \velocityRefHVKeep$),
\item $\policyVehicleAheadSlow$ - altruistic driver who slows down ($\velocityRefHV = \velocityRefHVSlow$).
\end{itemize}
In the intersection scenario we consider two types of driver:
\begin{itemize}
\item $\PolicyCross$ - driver keeps speed ($\velocityRefHV = \velocityRefHVKeep$),
\item $\PolicyStop$ - driver slows down to a stop.
\end{itemize}

\def\CTcross{CT_{\text{cross}}}
\def\CTstop{CT_{\text{stop}}}
\def\DistanceToIntersection{d}
\def\CruisingSpeed{v}
\def\MaxAcceleration{A_{\text{max}}}
\def\MaxBraking{D_{\text{max}}}
\def\TimeToContact{\text{TTC}}

\def\DistanceToIntersectionOncoming{d_o}
\def\CruisingSpeedOncoming{v_o}

At point $\sBranching$ the human decides on one of the policies to follow.
We make use of research in the field of neuroscience, namely we consider the work in~\cite{marti2015drivers}, where the authors study the decision making process of human drivers approaching an intersection.
The authors propose that drivers' decision making depends on the degree of safety of the two co-existing possibilities of either crossing or stopping at the intersection.
The degree of safety can be quantified by the critical time to cross $\CTcross$, and the critical time to stop $\CTstop$.
The probability of the human choosing to cross, \textit{i.e.}, applying policy $\PolicyCross$, is defined as~\cite{marti2015drivers}:
\begin{equation}
\label{\baselabel eq:crossing_frequency}
\ProbPolicyCross = \frac{w}{1+e^{-a \CTcross}} + \frac{1-w}{1+e^{-b \CTstop}}.
\end{equation}
Where the parameters $a$, $b$, and $w$ are found by fitting the model~\Cref{\baselabel eq:crossing_frequency} to experimental data.
The experimental data is obtained from thirty experienced drivers whose decision making is studied in a driving simulator.

\Cref{\baselabel eq:crossing_frequency} provides a decision making model that can be used to estimate the probabilities of the human driver taking different decisions.
This allows the AV to understand the likelihood of different future scene outcomes stemming from different decision taken by the human driver.
Moreover, the dependency of~\Cref{\baselabel eq:crossing_frequency} on both the HV and AV states allows the planner to reason about how different maneuvers affect the likelihood of different scenarios.
This is fundamental in order for the AV to achieve assertive behavior.
The following section proposes a framework that can take advantage of model~\Cref{\baselabel eq:crossing_frequency}, providing the planner with a better knowledge of HV behavior and allowing the AV to drive assertively.

\section{Motion Planning Framework}
\label{\baselabel sec:mpc_formulation}

\def\TreeFigureWidth{\columnwidth}

\subsection{Tree formulation - human-driven vehicle}

\begin{figure}
  \centering
	\resizebox {\TreeFigureWidth} {!} {
  	\begin{tikzpicture}[scale=0.75]
    \input{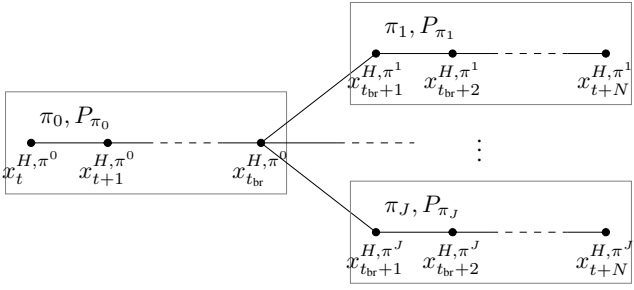}
	\end{tikzpicture}
	}
  \caption{Diagram of the scenario tree used to model the HV possible future states.}
  \label{fig:tree_diagram_human}
\end{figure}

In order to take into account the possible future decisions of the human, and the resulting states of its vehicle, we use a tree structure~\cite{chen2021interactive}.
\Cref{fig:tree_diagram_human} shows a tree with $J+1$ branches (branches $[2, \ldots, J-1]$ not visible).
Each branch $j$ in the tree has an associated human driver policy $\pi \PolicySuperOrSubScript j$.
Within a branch $j$, the HV states are propagated assuming they follow the associated policy $\pi \PolicySuperOrSubScript j$.
The evolution of states corresponds to a discretized model of~\Cref{\baselabel eq:vehicle_model} so that 
\begin{equation}
\label{\baselabel eq:discretized_human_model}
x_{t+1}^{H,\pi \PolicySuperOrSubScript j} =
f^{H,\pi \PolicySuperOrSubScript j} \left( x_{t}^{H,\pi \PolicySuperOrSubScript j}, x_{t}^{A,\pi \PolicySuperOrSubScript j} \right),
\end{equation}
where the vehicle input $u_{t}^{H,\pi \PolicySuperOrSubScript j}$ is determined according the active policy $\pi \PolicySuperOrSubScript {j}$, and can depend on the AV state, as in~\Cref{\baselabel eq:policy_velocity_following}.
For the transition states between branches
\begin{equation}
\label{\baselabel eq:discretized_human_model_branching_states}
x_{\branchingTimeIndex+1\GivenCurrentInformation}^{H,\pi \PolicySuperOrSubScript j} =
f^{H,\pi \PolicySuperOrSubScript j} \left( x_{\branchingTimeIndex\GivenCurrentInformation}^{H,\pi \PolicySuperOrSubScript 0}, x_{\branchingTimeIndex\GivenCurrentInformation}^{A,\pi \PolicySuperOrSubScript 0} \right),
\end{equation}
where $u_{\branchingTimeIndex\GivenCurrentInformation}^{H,\pi \PolicySuperOrSubScript j}$ already follows the policy $\pi \PolicySuperOrSubScript j$.

The tree in~\cref{fig:tree_diagram_human} is composed of a root branch and $J$ child branches, corresponding to the set of branches $\SetOfNodes = \{ 0, 1, 2, \ldots, J\}$.
The root branch splits into different branches at time $\branchingTimeIndex$, corresponding to the state at which the HV crosses the path length $\sBranching$.
At this path length $\sBranching$, the human changes to a policy that interacts with the AV.
Since the human can take multiple policies, $\branchingState\GivenCurrentInformation$ propagates into $J$ different states $x_{t_{br}+1\GivenCurrentInformation}^{H,\pi \PolicySuperOrSubScript j}$, corresponding to the different branches with policies $\pi \PolicySuperOrSubScript j$.
Each branch has an associated probability $P_{\pi \PolicySuperOrSubScript j}$.
For the first branch $P_{\pi \PolicySuperOrSubScript 0} = 1$, and for the remaining branches $\sum_{j=1}^{J} P_{\pi \PolicySuperOrSubScript j} = 1$.

\subsection{Tree formulation - autonomous vehicle}

\begin{figure}
  \centering
	\resizebox {\TreeFigureWidth} {!} {
  	\begin{tikzpicture}[scale=0.75]
    \input{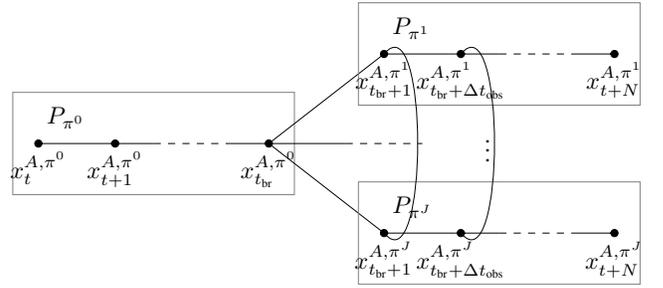}
	\end{tikzpicture}
	}
  \caption{Diagram of the scenario tree used to model the AV possible future states.
  The arcs between states in branch $1$ and branch $J$ correspond to autonomous vehicle states that are forced to be equal, according to \Cref{\baselabel eq:observation_constraints}.}
  \label{fig:tree_diagram_auto}
\end{figure}

The AV state evolution follows a similar tree structure as the HV, illustrated in~\cref{fig:tree_diagram_auto}, and corresponding to a discretized model of~\Cref{\baselabel eq:vehicle_model} so that
\begin{equation}
\label{\baselabel eq:discretized_autonomous_model}
x_{t+1\GivenCurrentInformation}^{A,\pi \PolicySuperOrSubScript j} =
f^{A,\pi \PolicySuperOrSubScript j} \left( x_{t\GivenCurrentInformation}^{A,\pi \PolicySuperOrSubScript j}, u_{t\GivenCurrentInformation}^{A,\pi \PolicySuperOrSubScript j} \right).
\end{equation}
The vehicle input $u_{i\GivenCurrentInformation}^{A,\pi \PolicySuperOrSubScript j}$ is determined by the solution to the optimal control problem introduced later in~\cref{\baselabel subsec:MPC_formulation}.
Similarly, for the branching states, we have:
\begin{equation}
\label{\baselabel eq:discretized_autonomous_model_branching_states}
x_{\branchingTimeIndex+1\GivenCurrentInformation}^{A,\pi \PolicySuperOrSubScript j} =
f^{A,\pi \PolicySuperOrSubScript j} \left( x_{\branchingTimeIndex\GivenCurrentInformation}^{A,\pi \PolicySuperOrSubScript 0}, u_{i\GivenCurrentInformation}^{A,\pi \PolicySuperOrSubScript j} \right).
\end{equation}

In a practical setting, it is not possible to immediately estimate the policy decision made by the human-driven vehicle, as prediction systems have a delay until accurately estimating the new human behavior.
Therefore, we force the autonomous vehicle states to be equal between the different branches $[1, \ldots, J]$ for the first $\ObservationTime$ seconds of the branch:
\def\breakline{\\}
\begin{multline}
\label{\baselabel eq:observation_constraints}
x_{t_i\GivenCurrentInformation}^{\RobotScript, \Policy{j}} = x_{t_i\GivenCurrentInformation}^{\RobotScript, \Policy{j'}}, \breakline \forall t_i \in [t_{\text{br}}+1, \ldots, \ObservationTimeInstant], \{ \forall j, j' \in \SetOfNodes | j \neq j' \}.
\end{multline}
\Cref{\baselabel eq:observation_constraints} forces the solutions to not assume immediate knowledge of the HV policy, and instead delay by $\ObservationTime$ the adaption to the new behavior.
$\ObservationTime$ is tuned based on considerations of expected time to perceive a new vehicle behavior and feasibility of the planning problem.

\subsection{MPC formulation}
\label{\baselabel subsec:MPC_formulation}

\def\VectorDefinitionTmpA{\NodeHVStateVector &= [ x_{t_i^j\GivenCurrentInformation}^{\HumanScript,\Policy{j}}, x_{t_i^j + 1\GivenCurrentInformation}^{\HumanScript,\Policy{j}}, \ldots, x_{t_f^j\GivenCurrentInformation}^{\HumanScript,\Policy{j}} ]}

\def\VectorDefinitionTmpB{\NodeAVStateVector &= [ x_{t_i^j\GivenCurrentInformation}^{\RobotScript,\Policy{j}}, x_{t_i^j + 1\GivenCurrentInformation}^{\RobotScript,\Policy{j}}, \ldots, x_{t_f^j\GivenCurrentInformation}^{\RobotScript,\Policy{j}} ]}

\def\VectorDefinitionTmpC{\NodeAVInputVector &= [ u_{t_i^j\GivenCurrentInformation}^{\RobotScript,\Policy{j}}, u_{t_i^j + 1\GivenCurrentInformation}^{\RobotScript,\Policy{j}}, \ldots, u_{t_f^j\GivenCurrentInformation}^{\RobotScript,\Policy{j}} ]}

For each branch $j \in \SetOfNodes$, consider the vectors
\begin{align*}
\VectorDefinitionTmpA ,\\
\VectorDefinitionTmpB ,\\
\VectorDefinitionTmpC ,
\end{align*}
where $t_i^j$ and $t_f^j$ are initial and final times associated with the first and last states in branch $j$.
$\NodeHVStateVector$ corresponds to the HV states, and $\NodeAVStateVector$, $\NodeAVInputVector$ to the AV states, and inputs, respectively.

\def\OptSpace{ ~ }

\def\CntHStates{
\mathrm{\cref{\baselabel eq:discretized_human_model}}, \forall i = t_i^j, t_i^j +1, \ldots, t_f^j, \forall j \in \SetOfNodes
}
\def\CntAStates{
\mathrm{\cref{\baselabel eq:discretized_autonomous_model}}, \forall i = t_i^j, t_i^j + 1, \ldots, t_f^j, \forall j \in \SetOfNodes
}

\def\CntHBranchingStates{
\mathrm{\cref{\baselabel eq:discretized_human_model_branching_states}}, \forall j \in \SetOfNodesMinusRoot
}
\def\CntABranchingStates{
\mathrm{\cref{\baselabel eq:discretized_autonomous_model_branching_states}}, \forall j \in \SetOfNodesMinusRoot
}

\def\CntObsStates{
\mathrm{\cref{\baselabel eq:observation_constraints}}, \forall j \in \SetOfNodesMinusRoot
}

\def\CntNoisyRationalLeafNodes{
\mathrm{\cref{\baselabel eq:noisy_rational}},\; j \in \SetOfNodesMinusRoot
}
\def\CntPsychologyProbabilityLeafNodes{
\mathrm{\cref{\baselabel eq:crossing_frequency}},\; j \in \SetOfNodesMinusRoot
}

\def\CntNoisyRationalRoot{
P_{\pi \PolicySuperOrSubScript 0} = 1
}

\def\CntInitialStates{
\begin{cases}
x_{t\GivenCurrentInformation}^{H, \pi \PolicySuperOrSubScript 0} \shorteq \InitialStateHV , x_{t\GivenCurrentInformation}^{A, \pi \PolicySuperOrSubScript 0} \shorteq \InitialStateAV  & \text{if } t \leq t_{\text{br}}\\
x_{t\GivenCurrentInformation}^{H, \pi \PolicySuperOrSubScript j} \shorteq \InitialStateHV , x_{t\GivenCurrentInformation}^{A, \pi \PolicySuperOrSubScript j} \shorteq \InitialStateAV, j \in \SetOfNodesMinusRoot & \text{otherwise}
\end{cases}
}

\def\CntProbabilityRoot{
\begin{cases}
P_{\pi \PolicySuperOrSubScript 0} = 1   & \text{if } t \leq t_{\text{br}}\\
P_{\pi \PolicySuperOrSubScript 0} = 0   & \text{otherwise}
\end{cases}
}

\def\CntPsychologyProbabilityLeafNodesNew{
\begin{cases}
\mathrm{\cref{\baselabel eq:crossing_frequency}},\; j \in \SetOfNodesMinusRoot   & \text{if } t \leq \ObservationTimeInstant \\
P_{\pi \PolicySuperOrSubScript {\ActualNode}} \shorteq 1, P_{\pi \PolicySuperOrSubScript {j}} \shorteq 0,\; j \in \SetOfNodesMinusArgument{\ActualNode}   & \text{otherwise}
\end{cases}
}

The MPC is solved at the beginning of every planning cycle at time $t$, its formulation is as follows:
\begin{subequations}
	\allowdisplaybreaks 
	\label{\baselabel eq:optimization_problem}
	\begin{align} 
	\underset{ \{ \NodeAVInputVector \}_{j \in \SetOfNodes} }{\text{min}} {\hskip0.25em\relax} & \sum_{j \in \SetOfNodes} \OptSpace P_{\pi \PolicySuperOrSubScript j} J ( \NodeHVStateVector, \NodeAVStateVector, \NodeAVInputVector ) \label{\baselabel eq:optimization_objective}\\
	\text{s.t.} {\hskip1.0em\relax} & \CntHStates \label{\baselabel eq:constraint_1}\\
	& \CntHBranchingStates , \label{\baselabel eq:constraint_3}\\
	& \CntAStates , \label{\baselabel eq:constraint_2}\\
	& \CntABranchingStates , \label{\baselabel eq:constraint_4}\\
	& \CntObsStates , \label{\baselabel eq:constraint_5}\\
	& \CntProbabilityRoot , \label{\baselabel eq:cnt_prob_root_node}\\
	& \CntPsychologyProbabilityLeafNodesNew \label{\baselabel eq:cnt_prob_leaf_nodes}\\
	& \CntInitialStates \label{\baselabel eq:cnt_initial_states}
	\end{align} 
\end{subequations}

\Cref{\baselabel eq:constraint_1,\baselabel eq:constraint_2} encode the state evolution constraints within the branches for the HV and AV.
Similarly, \Cref{\baselabel eq:constraint_3,\baselabel eq:constraint_4} encode the state evolution constraints between branches.
\Cref{\baselabel eq:constraint_5} sets the observation constraints forcing the AV states to be equal across branches $j \in \SetOfNodesMinusRoot$, from time $t_{\text{br}}$ until time $\ObservationTimeInstant$.
\Cref{\baselabel eq:cnt_prob_root_node} defines the probability of the root branch: in case the current planning cycle happens before the HV has taken a decision, \textit{i.e.}, $t \leq t_{\text{br}}$, the branch exists and is certain, $P_{\pi \PolicySuperOrSubScript 0} = 1$.
Otherwise, the branch does not exist anymore, and it is disregarded in the optimization objective, $P_{\pi \PolicySuperOrSubScript 0} = 0$.
\Cref{\baselabel eq:cnt_prob_leaf_nodes} defines the probabilities of the leaf branches: in case the current planning cycle happens before the AV can identify the true mode of the human, \textit{i.e.}, $t \leq \ObservationTimeInstant$, its probability is given by the behavioral decision making model in~\cite{marti2015drivers}.
Otherwise, the probability of the branch corresponding to the true mode is set to 1, and the probability of all other branches is 0, and they are ignored from~\Cref{\baselabel eq:optimization_problem}.
Finally,~\Cref{\baselabel eq:cnt_initial_states} makes the initial HV and AV states of the root branch correspond to the current measured states $\InitialStateHV, \InitialStateAV$ of the vehicles if $t \leq t_{\text{br}}$.
Otherwise it makes the initial state of all possible branches equal to $\InitialStateHV, \InitialStateAV$.

Solving optimization problem~\Cref{\baselabel eq:optimization_problem}, we find the optimal AV input vectors $\NodeAVInputVectorOptimal$ for each branch of the tree.
At each planning cycle, the optimization problem~\Cref{\baselabel eq:optimization_problem} is solved, considering the latest sensor measurements $\InitialStateHV, \InitialStateAV$.
The vehicle control input corresponds to the first element of the root branch input vector $\NodeAVInputVectorOptimal[0]$ if $t \leq t_{\text{br}}$.
Otherwise, it corresponds to the first element of an active child branch.

\section{Results}
\label{\baselabel sec:results}

The following results are obtained in a custom Python simulation environment.
The MPC problem presented in~\Cref{\baselabel eq:optimization_problem} is formulated using CasADi~\cite{andersson2019casadi}, and the corresponding nonlinear optimization problem is solved using Ipopt~\cite{wachter2006implementation}.

\subsection{Uncertain Traffic Light}

\begin{figure}
\centering
\includegraphics[width=0.95\columnwidth]{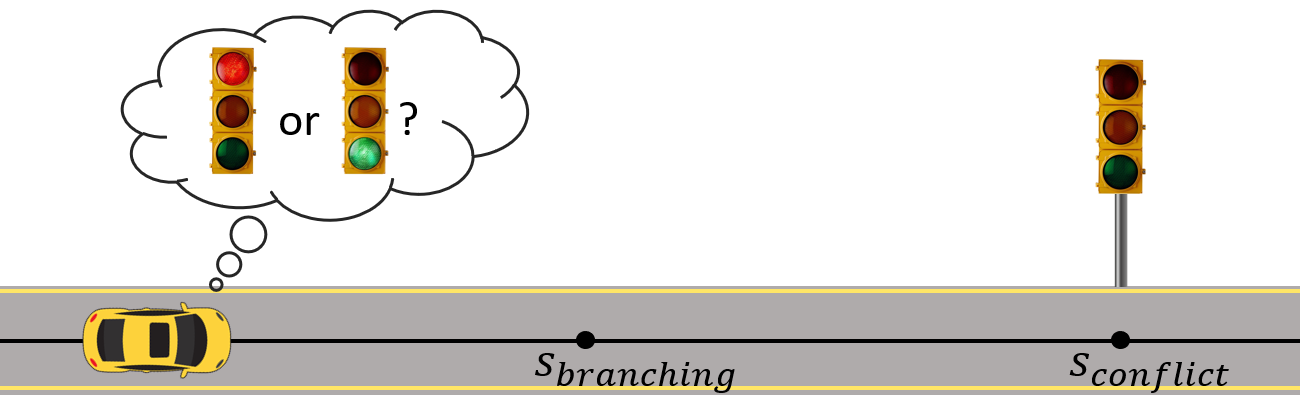}
\caption{Scenario with an uncertain traffic light.}
\label{\baselabel fig:uncertain_obstacle_scenario}
\end{figure}

We present the results of a scenario where a traffic light on the road cannot be adequately perceived by the AV sensors, illustrated in~\cref{\baselabel fig:uncertain_obstacle_scenario}.
The AV knows about the existence of the traffic light, due to information contained in maps or since the perception systems detect the existence of the traffic light pole.
However, it does not know the state of the traffic light due to poor visibility conditions.
We assume that at $\sBranching$, the AV accurately detects the state of the traffic light.

The tree structure for this scenario corresponds to a root node lasting until the AV crosses $\sBranching$, with two child nodes, one that continues driving if the traffic light is green, $\PolicyGreenLight$, and one coming to a stop if the traffic light is red, $\PolicyRedLight$.

\begin{figure}
\centering
\includegraphics[width=1.0\columnwidth]{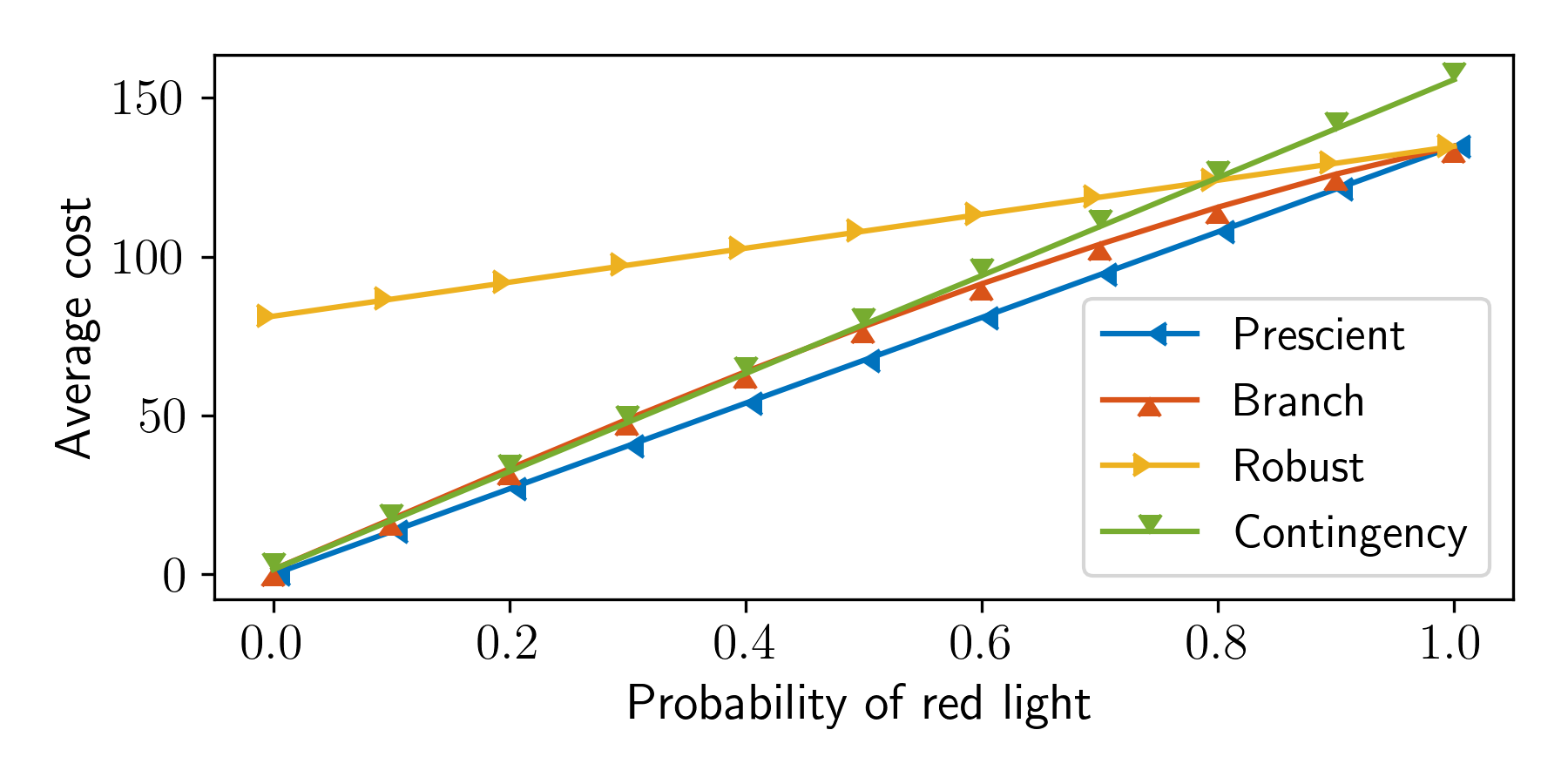}
\caption{Average closed-loop costs of Robust, Prescient, Contingency, and Branch MPC for different probabilities of the scenario's traffic light being red.}
\label{\baselabel fig:uncertain_obstacles_comparison_average}
\end{figure}

We study the performances of a Robust, Prescient, and Contingency MPC and compare them to our proposed Branch MPC.
The different formulations are as follows:
\begin{itemize}
\item Robust MPC - assumes the traffic light is red until $\sBranching$, afterward it detects the actual traffic light state;
\item Prescient MPC - assumes a perfect perception system that knows the state of the traffic light even before $\sBranching$;
\item Contingency MPC - optimizing the driving behavior as if the traffic light was green, but keeping a contingency plan if the light is red;
\item Branch MPC - our approach, considering both traffic light possibilities and assuming $\ProbPolicyGreenLight = \ProbPolicyRedLight = 0.5$.
\end{itemize}
We run closed-loop simulations for different probabilities of the traffic light being red.
The average performance of the different approaches is shown in~\cref{\baselabel fig:uncertain_obstacles_comparison_average}.

The Robust MPC has an expected cost that is never better than the Prescient or Branch MPCs.
Since Robust MPC always assumes that the obstacle exists, it achieves an acceptable performance when the probability of the obstacle is high enough.
However, in most cases, it has the worst performance.
On the other hand, the Prescient MPC consistently achieves the best performance as it plans its maneuver according to the actual (unknown) state of the traffic light.

The Contingency MPC performance is comparable to the Branch and Prescient MPCs for small obstacle probabilities.
However, as the obstacle probability increases, its performance degrades.
The Contingency MPC is an optimistic planner, assuming a green light until proven otherwise.
This results in abrupt braking maneuvers when the light is red, leading to poor performance.

Our proposed method achieves, on average, a performance that is always better than Robust MPC while being equally as safe.
For lower probabilities of red light, the Branch MPC performance is comparable to that of Contingency MPC.
For higher probabilities, Branch MPC outperforms Contingency MPC due to not being optimistic about the traffic light state.

\begin{figure}
\centering
\includegraphics[width=0.95\columnwidth]{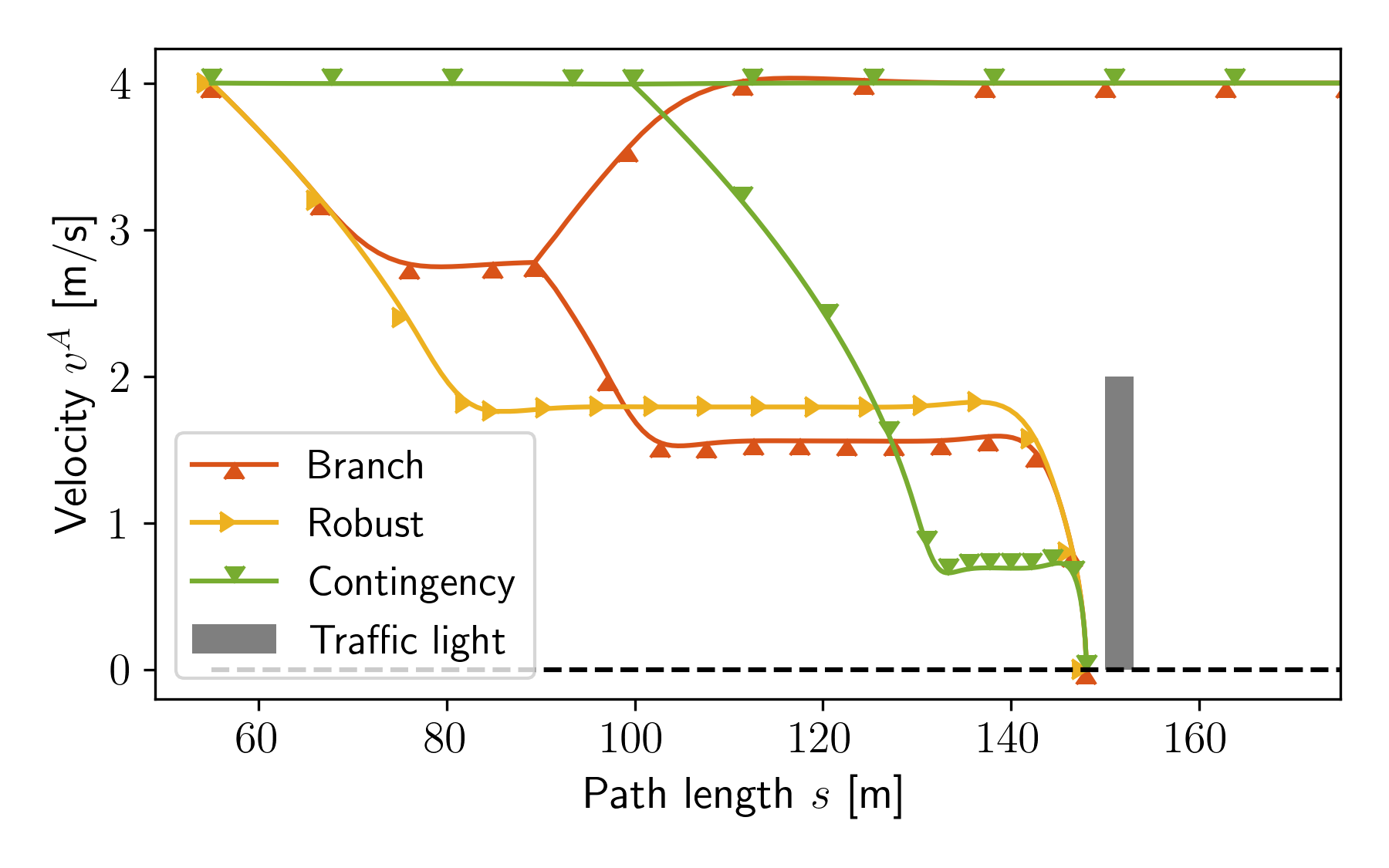}
\caption{Single planning cycle for the autonomous vehicle approaching an uncertain obstacle under different planning frameworks: Branch, Robust, and Contingency MPCs.}
\label{\baselabel fig:uncertain_obstacle_single_planning}
\end{figure}

\Cref{\baselabel fig:uncertain_obstacle_single_planning} shows a single planning cycle for the different MPC approaches starting at identical initial vehicle states.
In this initial state, the AV is aware of the traffic light but does not know its state yet.
Robust MPC plans a single velocity profile assuming a red light and bringing the vehicle to a stop.
Contingency MPC plans two velocity profiles to deal with the possibility of green and red lights.
The velocity profile associated with a red light is not penalized in the cost function, resulting in an abrupt and uncomfortable contingency maneuver.
Branch MPC reduces its driving speed, at the cost of increasing travel time, to deal with the possibility of a red light comfortably.
Therefore, Branch MPC chooses a tradeoff between the performances in both future possible scenarios.

\subsection{Multi-modality}
\label{subsec:multimodality}

We consider the merging scenario in~\cref{\baselabel fig:y_merging_scenario}, where the HV can have three possible future velocity tracking policies $\policyVehicleAheadFast$, $\policyVehicleAheadKeep$, or $\policyVehicleAheadSlow$.
The tree structure consists of a root node where the human driver keeps a constant speed policy $\pi_0$ until the branching point, and three child nodes corresponding to the HV taking the three possible policies.
Although chosen manually in this study, the tracking policies could be given by a prediction module providing a multi-modal set of outcomes, as in~\cite{chai2020multipath,ivanovic2021MATS}.

\begin{figure}
\centering
\includegraphics[width=0.95\columnwidth]{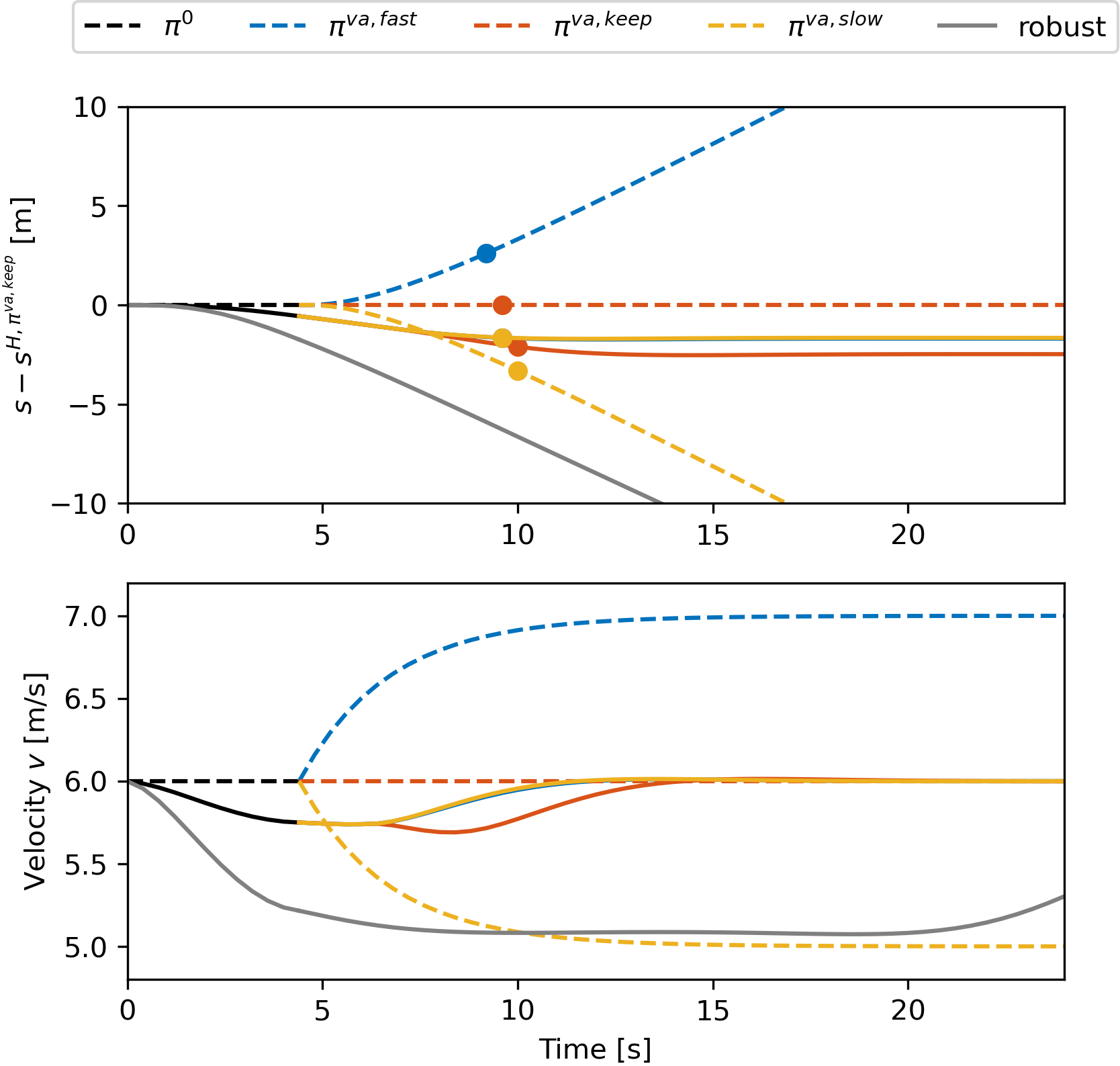}
\caption{Planned autonomous vehicle trajectory (solid line) and predicted human-driven vehicle trajectory (dashed line) for the Branch MPC and Robust MPC cases. Top: Path lengths (centered around the path executed by $\policyVehicleAheadKeep$). The rounded markers correspond to the instant when the vehicles cross the conflict point. Bottom: Velocity profiles.}
\label{\baselabel fig:merge_scenario_single_plan_tree}
\end{figure}

\Cref{\baselabel fig:merge_scenario_single_plan_tree} (bottom) shows the planned velocity profiles of a planning cycle occurring when the HV and AV are heading toward the merging point.
The HV is predicted to have the possibility to take either of the three different velocity profiles.
The AV plans a set of velocity profiles, going behind the HV in case it decides to keep its speed or accelerate or going ahead in case the HV slows down.
The AV plan is better visualized by looking at the planned path lengths in~\cref{\baselabel fig:merge_scenario_single_plan_tree} (top).
The path lengths (subtracted by the predicted path length of the HV with policy $\policyVehicleAheadKeep$) show the distinctive decisions made by the planner to either go behind or ahead of different possible policies of the HV.

\Cref{\baselabel fig:merge_scenario_single_plan_tree} also shows the velocity planned by a Robust MPC approach.
The planned maneuver is conservative, opting to go behind all possible realizations of future human policies.
Considering the multi-modality and planning feedback policies dependent on the mode that the human eventually decides upon allows the planned profiles to squeeze in between different possible policies, reducing conservativeness compared to a robust approach.

\subsection{Non-interaction \textit{vs.} interaction-aware human models}

We now consider an AV that accelerates from a standstill and merges onto a road with an oncoming vehicle driving at high speed.
The AV can decide to go ahead of the other vehicle or wait for it to pass and go behind it.

\begin{figure}
\centering
\includegraphics[width=\columnwidth]{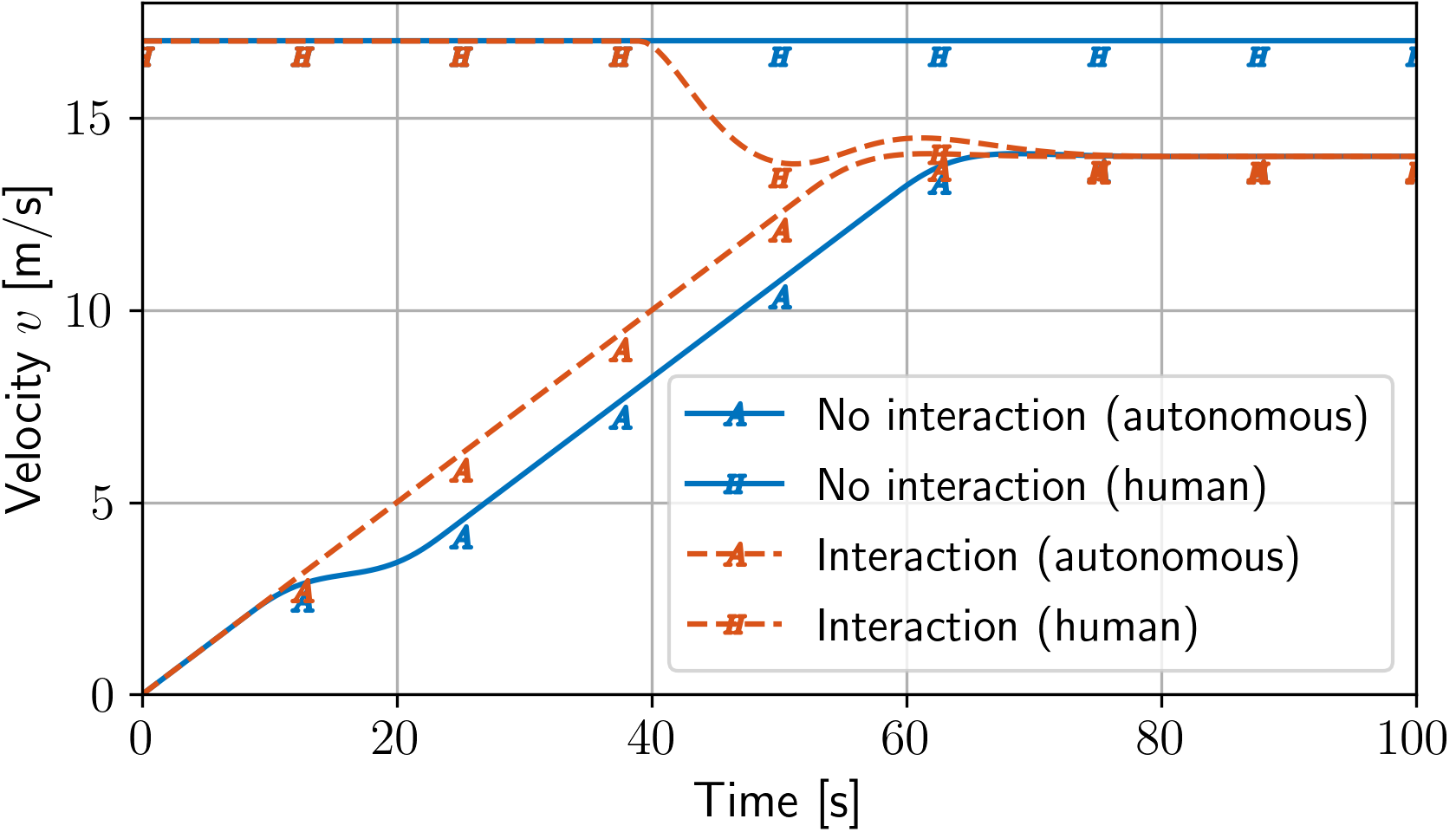}
\caption{Predicted and planned velocity profiles when considering non-interacting and interaction-aware HV models.}
\label{fig:interaction_aware_comparison}
\end{figure}

\Cref{fig:interaction_aware_comparison} shows a single planning cycle, where the AV is at a standstill, and an oncoming vehicle drives faster than the AV's desired cruising velocity.
In the non-interacting vehicle model case, the AV predicts the HV to keep its speed constant and decides to wait for it to pass and, afterward, go behind it.
In the case of packed traffic with several oncoming vehicles, the AV could be stuck indefinitely at the junction~\cite{trautman2010unfreezing}.
However, when considering an interaction-aware model, the AV decides to go ahead of the oncoming vehicle, as it predicts that the HV will slow down and adapt its speed.
We remark that the planner assumes a limit to the HV's braking capabilities to guarantee that the AV does not act discourteously to other traffic participants~\cite{sun2018courteous}.

\subsection{Intersection scenario}

We consider the non-signalized intersection scenario in~\cref{\baselabel fig:intersection_scenario}.
Since there are no priority rules, the vehicles must negotiate who goes first.
The tree structure consists of a root node and two child nodes corresponding to the HV keeping its speed, $\PolicyCross$, or stopping, $\PolicyStop$.

\begin{figure}
    \centering
    \begin{subfigure}[b]{\columnwidth}
        \includegraphics[width=\textwidth]{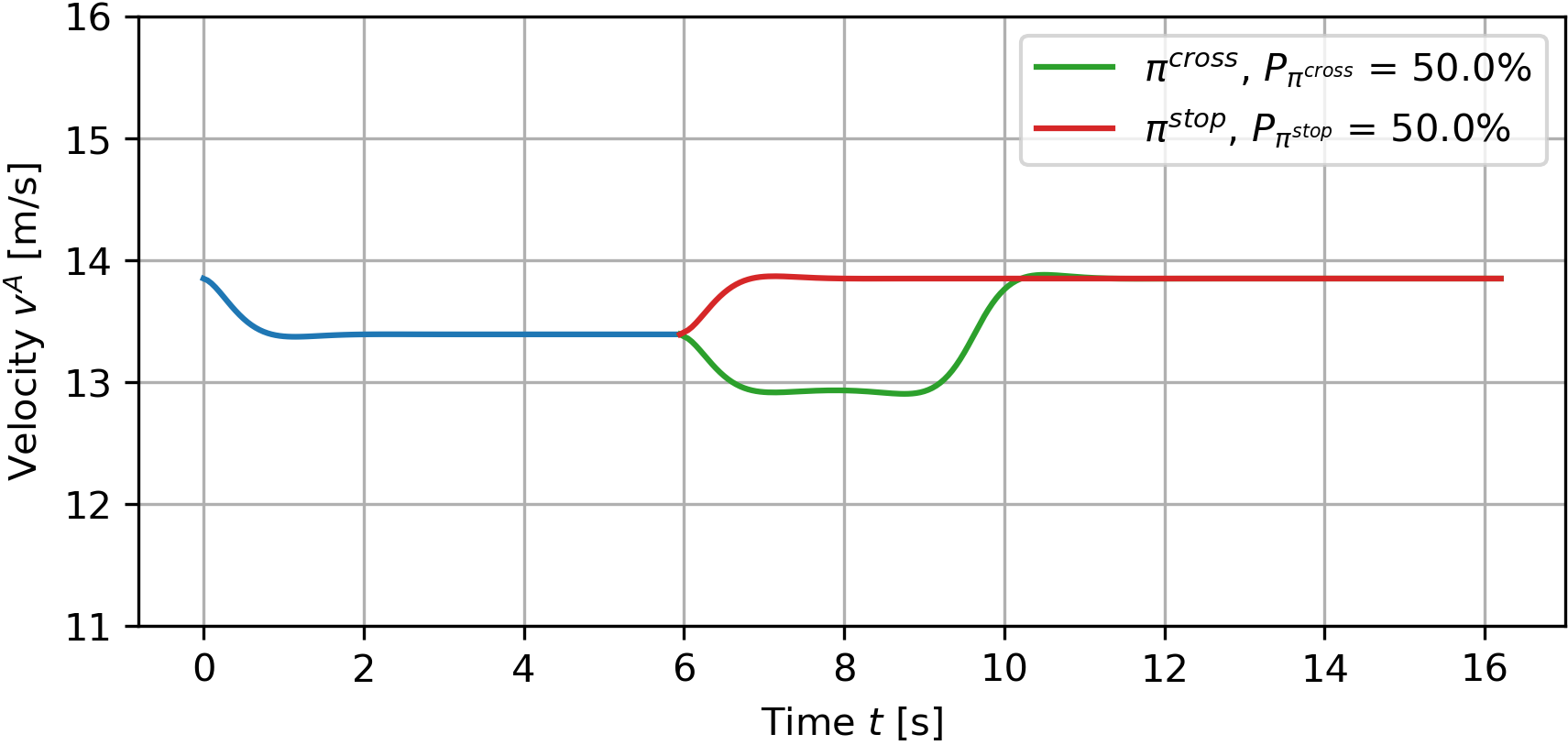}
        \caption{Fixed probabilities.}
        \label{\baselabel fig:intersection_fixed_probabilities}
    \end{subfigure}
      
    \begin{subfigure}[b]{\columnwidth}
        \includegraphics[width=\textwidth]{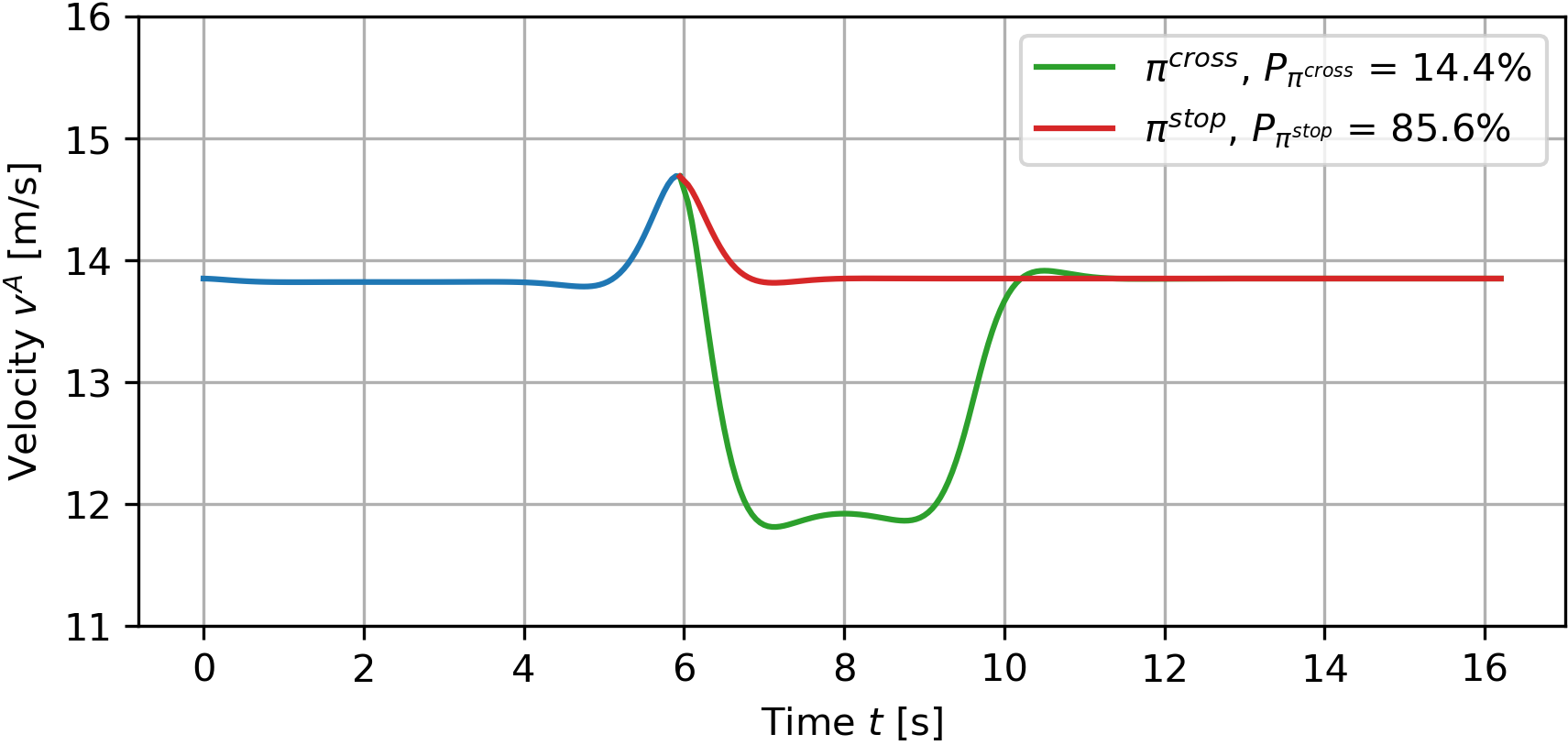}
        \caption{Adapting probabilities.}
        \label{\baselabel fig:intersection_varying_probabilities}
    \end{subfigure}
    \caption{Intersection scenario for different configurations of human-vehicle policy probabilities.}\label{\baselabel fig:intersection_scenario_probabilities}
\end{figure}

\Cref{\baselabel fig:intersection_fixed_probabilities} shows the AV planned velocities when assuming that both HV policies have an equal and fixed probability $\ProbPolicyCross = \ProbPolicyStop = 0.5$.
The AV decides to slow down to deal with both possible outcomes of the human decision.

We note that this intersection scenario resembles the one considered in~\cite{marti2015drivers}, and therefore we use the human decision making model~\Cref{\baselabel eq:crossing_frequency} to determine probabilities $\ProbPolicyCross$ and $\ProbPolicyStop$.
\Cref{\baselabel fig:intersection_varying_probabilities} shows the results for this scenario when considering that the human policy probabilities follow~\Cref{\baselabel eq:crossing_frequency}.
In this case, the planned velocity profile increases at around $t = 5$ s to assertively indicate to the human driver that the AV intends to cross.
This results in a higher predicted probability of the human deciding to slow down and give way to the autonomous vehicle, $\ProbPolicyStop = 0.856$.
With a lower probability of the human keeping its speed, $\ProbPolicyCross = 0.144$, the AV plans a more abrupt maneuver in the unlikely event of this outcome occurring.

These results show that considering the human decision making model~\Cref{\baselabel eq:crossing_frequency} allows the planner to make assertive maneuvers and improve driving performance.
The AV drives in a way that shows intent to other HVs, intending to optimize its expected driving outcomes.
This behavior comes out naturally due to minimizing the objective cost of~\Cref{\baselabel eq:optimization_objective}, and therefore, does not require manually tuned driving strategies for different traffic situations.

\section{Conclusions}
\label{\baselabel sec:conclusions}

We presented a motion planning framework tackling the challenges of autonomous vehicles in the presence of human drivers, namely, multi-modality, interaction, and decision making.
A Branch MPC problem is formulated by combining scenario trees and research from the neuroscience field to model the human driver's decision making process.
We show that Branch MPC and interaction-aware models achieve better average performance than alternative formulations in the literature.
Furthermore, we show that using a human decision making model leads the planner to find proactive maneuvers that convey intent to the human driver.
The proposed framework plans assertive maneuvers that influence human drivers to make decisions favorable to the autonomous vehicle.

In future work, it is interesting to analyze the impact of the problem modeling assumptions and to consider behavioral decision making for more general driving scenarios.
Finally, practical implementation is needed to validate the suitability of the proposed approach to model and solve the joint prediction and planning problem accurately.

\bibliographystyle{ieeetran}
\bibliography{references}

\begin{thebibliography}{10}
\providecommand{\url}[1]{#1}
\csname url@samestyle\endcsname
\providecommand{\newblock}{\relax}
\providecommand{\bibinfo}[2]{#2}
\providecommand{\BIBentrySTDinterwordspacing}{\spaceskip=0pt\relax}
\providecommand{\BIBentryALTinterwordstretchfactor}{4}
\providecommand{\BIBentryALTinterwordspacing}{\spaceskip=\fontdimen2\font plus
\BIBentryALTinterwordstretchfactor\fontdimen3\font minus
  \fontdimen4\font\relax}
\providecommand{\BIBforeignlanguage}[2]{{%
\expandafter\ifx\csname l@#1\endcsname\relax
\typeout{** WARNING: IEEEtran.bst: No hyphenation pattern has been}%
\typeout{** loaded for the language `#1'. Using the pattern for}%
\typeout{** the default language instead.}%
\else
\language=\csname l@#1\endcsname
\fi
#2}}
\providecommand{\BIBdecl}{\relax}
\BIBdecl

\bibitem{batkovic2020robust}
I.~Batkovic, U.~Rosolia, M.~Zanon, and P.~Falcone, ``A robust scenario mpc
  approach for uncertain multi-modal obstacles,'' \emph{IEEE Control Systems
  Letters}, vol.~5, no.~3, pp. 947--952, 2020.

\bibitem{phiquepal2021control}
C.~Phiquepal and M.~Toussaint, ``Control-tree optimization: an approach to mpc
  under discrete partial observability,'' in \emph{2021 IEEE International
  Conference on Robotics and Automation (ICRA)}.\hskip 1em plus 0.5em minus
  0.4em\relax IEEE, 2021, pp. 9666--9672.

\bibitem{schildbach2015scenario}
G.~Schildbach and F.~Borrelli, ``Scenario model predictive control for lane
  change assistance on highways,'' in \emph{2015 IEEE Intelligent Vehicles
  Symposium (IV)}, 2015, pp. 611--616.

\bibitem{cesari2017scenario}
G.~Cesari, G.~Schildbach, A.~Carvalho, and F.~Borrelli, ``Scenario model
  predictive control for lane change assistance and autonomous driving on
  highways,'' \emph{IEEE Intelligent Transportation Systems Magazine}, vol.~9,
  no.~3, pp. 23--35, 2017.

\bibitem{brudigam2018scsmpc}
T.~Brüdigam, M.~Olbrich, M.~Leibold, and D.~Wollherr, ``Combining stochastic
  and scenario model predictive control to handle target vehicle uncertainty in
  an autonomous driving highway scenario,'' in \emph{2018 21st International
  Conference on Intelligent Transportation Systems (ITSC)}, 2018, pp.
  1317--1324.

\bibitem{alsterda2019contingency}
J.~P. Alsterda, M.~Brown, and J.~C. Gerdes, ``Contingency model predictive
  control for automated vehicles,'' in \emph{2019 American Control Conference
  (ACC)}, 2019, pp. 717--722.

\bibitem{alsterda2021contingency}
J.~P. Alsterda and J.~C. Gerdes, ``Contingency model predictive control for
  linear time-varying systems,'' \emph{arXiv preprint arXiv:2102.12045}, 2021.

\bibitem{nair2022interaction}
S.~H. Nair, V.~Govindarajan, T.~Lin, Y.~Wang, E.~H. Tseng, and F.~Borrelli,
  ``Stochastic mpc with dual control for autonomous driving with multi-modal
  interaction-aware predictions,'' \emph{arXiv preprint arXiv:2208.03525},
  2022.

\bibitem{sadigh2018planning}
D.~Sadigh, N.~Landolfi, S.~S. Sastry, S.~A. Seshia, and A.~D. Dragan,
  ``Planning for cars that coordinate with people: leveraging effects on human
  actions for planning and active information gathering over human internal
  state,'' \emph{Autonomous Robots}, vol.~42, no.~7, pp. 1405--1426, 2018.

\bibitem{marti2015drivers}
G.~Marti, A.~H. Morice, and G.~Montagne, ``Drivers' decision-making when
  attempting to cross an intersection results from choice between
  affordances,'' \emph{Frontiers in human neuroscience}, vol.~8, p. 1026, 2015.

\bibitem{chen2021interactive}
Y.~Chen, U.~Rosolia, W.~Ubellacker, N.~Csomay-Shanklin, and A.~D. Ames,
  ``Interactive multi-modal motion planning with branch model predictive
  control,'' \emph{IEEE Robotics and Automation Letters}, vol.~7, no.~2, pp.
  5365--5372, 2022.

\bibitem{trautman2010unfreezing}
P.~Trautman and A.~Krause, ``Unfreezing the robot: Navigation in dense,
  interacting crowds,'' in \emph{2010 IEEE/RSJ International Conference on
  Intelligent Robots and Systems}, 2010, pp. 797--803.

\bibitem{wang2022interaction}
R.~Wang, M.~Schuurmans, and P.~Patrinos, ``Interaction-aware model predictive
  control for autonomous driving,'' \emph{arXiv preprint arXiv:2211.17053},
  2022.

\bibitem{schwarting2019social}
W.~Schwarting, A.~Pierson, J.~Alonso-Mora, S.~Karaman, and D.~Rus, ``Social
  behavior for autonomous vehicles,'' \emph{Proceedings of the National Academy
  of Sciences}, vol. 116, no.~50, pp. 24\,972--24\,978, 2019.

\bibitem{treiber2000congested}
M.~Treiber, A.~Hennecke, and D.~Helbing, ``Congested traffic states in
  empirical observations and microscopic simulations,'' \emph{Physical review
  E}, vol.~62, no.~2, p. 1805, 2000.

\bibitem{andersson2019casadi}
J.~A. Andersson, J.~Gillis, G.~Horn, J.~B. Rawlings, and M.~Diehl, ``Casadi: a
  software framework for nonlinear optimization and optimal control,''
  \emph{Mathematical Programming Computation}, vol.~11, no.~1, pp. 1--36, 2019.

\bibitem{wachter2006implementation}
A.~W{\"a}chter and L.~T. Biegler, ``On the implementation of an interior-point
  filter line-search algorithm for large-scale nonlinear programming,''
  \emph{Mathematical programming}, vol. 106, no.~1, pp. 25--57, 2006.

\bibitem{chai2020multipath}
Y.~Chai, B.~Sapp, M.~Bansal, and D.~Anguelov, ``Multipath: Multiple
  probabilistic anchor trajectory hypotheses for behavior prediction,'' in
  \emph{Proceedings of the Conference on Robot Learning}, vol. 100.\hskip 1em
  plus 0.5em minus 0.4em\relax PMLR, 30 Oct--01 Nov 2020, pp. 86--99.

\bibitem{ivanovic2021MATS}
B.~Ivanovic, A.~Elhafsi, G.~Rosman, A.~Gaidon, and M.~Pavone, ``{MATS}: An
  interpretable trajectory forecasting representation for planning and
  control,'' in \emph{Proceedings of the 2020 Conference on Robot Learning},
  vol. 155.\hskip 1em plus 0.5em minus 0.4em\relax PMLR, 16--18 Nov 2021, pp.
  2243--2256.

\bibitem{sun2018courteous}
L.~Sun, W.~Zhan, M.~Tomizuka, and A.~D. Dragan, ``Courteous autonomous cars,''
  in \emph{2018 IEEE/RSJ International Conference on Intelligent Robots and
  Systems (IROS)}.\hskip 1em plus 0.5em minus 0.4em\relax IEEE, 2018, pp.
  663--670.

\end{thebibliography}


\end{document}